\documentclass[conference,letterpaper]{IEEEtran}

\usepackage[T1]{fontenc}
\usepackage{cite}
\usepackage{amsmath,amssymb,amsfonts}
\usepackage{graphicx}
\usepackage{textcomp}
\usepackage{xcolor}
\usepackage{url}
\usepackage{booktabs}
\usepackage{multirow}
\usepackage{tabularx}
\usepackage{flushend}
\usepackage[hidelinks]{hyperref}

\newcommand{\oracleA}{Oracle-A}
\newcommand{\oracleB}{Oracle-B}
\newcommand{\oracleAB}{Oracle-A+B}

\begin{document}

\title{What Breaks in Multi-Image Schema Application? Factorial Oracle Interventions Reveal an Instance-Information Bottleneck}
\date{}

\author{
\IEEEauthorblockN{Yiliu Xu\textsuperscript{*}}
\IEEEauthorblockA{Carnegie Mellon University\\
Pittsburgh, United States\\
yiliux@alumni.cmu.edu}
\and
\IEEEauthorblockN{Xinjin Li\textsuperscript{*}}
\IEEEauthorblockA{Columbia University\\
New York, United States\\
li.xinjin@columbia.edu}
\and
\IEEEauthorblockN{Yu Ma}
\IEEEauthorblockA{Carnegie Mellon University\\
Pittsburgh, United States\\
yuma13926@gmail.com}
}

\maketitle

\begingroup
\renewcommand{\thefootnote}{*}
\footnotetext{Yiliu Xu and Xinjin Li contributed equally to this work.}
\endgroup

\begin{abstract}
Modern vision-language models can accept multi-image contexts, but aggregate accuracy rarely reveals where a multi-image reasoning pipeline fails. We study a controlled schema-application task in which a model must locate two target characters in a 16-image sequence, identify their visual attributes, consult an item-specific visual codebook, and compose two ordered nonce code names. We introduce a within-item $2\times2$ factorial Oracle design: Oracle-A supplies a verified textual schema, Oracle-B supplies a verified ordered target-appearance representation, and both interventions retain the original images and final question. Across 72 new items and 1,188 analyzed model responses, Oracle-B increased strict exact accuracy by 54.6 percentage points (95\% confidence interval: 46.3--63.0) and outperformed Oracle-A by 60.6 points ($p_{\mathrm{Holm}}\leq .0002$). An auxiliary retrieval diagnostic reached only 49.1\% ordered-appearance accuracy even after removing code-name generation. Providing both Oracle sources yielded 100\% accuracy across all models. These results localize the dominant recoverable deficit to upstream ordered instance recovery rather than primarily to downstream schema use and composition.
\end{abstract}


\section{Introduction}

Large vision-language models (VLMs) are now routinely asked to handle multiple images and interleaved modalities. The benchmark picture is already clear: good single-image performance does not carry over cleanly to multi-image reasoning \cite{yue2024mmmu,zhao2024mirb,wang2025muirbench}, and accepting many visual tokens is not the same as finding the few that matter \cite{wang2025mmneedle,wang2024mmniah,wu2025visualhaystacks}. What remains less clear is the point of failure. A wrong final answer can hide very different mistakes.

The research problem is failure localization in a black-box multi-image pipeline. Given an incorrect final response, we seek to determine whether the dominant recoverable deficit lies in visual-schema acquisition, ordered target-instance recovery, or downstream lookup and composition. End-to-end accuracy cannot distinguish these possibilities because it collapses every stage into one outcome.

Our goal is to test which verified intermediate representation is sufficient to restore an otherwise unchanged task. We use a within-item $2\times2$ factorial Oracle design. Each synthetic item contains a visual schema (Evidence A), a dual-target scene (Evidence B), and 14 distractor images. The four conditions keep the same 16 images and question while adding either a verified schema transcript, verified ordered target appearances, both, or neither. The 16-image task is a controlled testbed for the diagnostic design rather than the contribution by itself.

If the bottleneck sits in the visual codebook, Oracle-A should help. If it sits in recovery of the two ordered target appearances, Oracle-B should help. Across three commercial VLMs, the result is strongly asymmetric: ordered instance information produces a 54.6-point strict rescue, while complete schema transcription barely changes semantic accuracy.

The contribution is methodological and empirical. First, the matched $\oracleA\times\oracleB$ design estimates the independent and joint effects of repairing two candidate intermediate representations without changing the visual context or final answer. Second, a balanced 72-item testbed and auxiliary retrieval controls connect the factorial contrast to representation-level behavior. Third, the experiments localize the dominant recoverable deficit to ordered instance recovery rather than downstream schema use and composition.

Section II positions the study relative to multi-image evaluation, long-context retrieval, and Oracle diagnostics. Section III defines the diagnostic design and its task instantiation. Section IV describes the evaluation protocol and statistical analysis. Sections V and VI report and interpret the results, followed by limitations and conclusions in Sections VII and VIII.

\section{Related Work}

This section reviews the three areas that directly motivate the study: multi-image evaluation, evidence use in long contexts, and Oracle-based failure diagnosis.

\subsection{Multi-Image VLM Evaluation}

General multimodal benchmarks document substantial remaining gaps in expert perception and reasoning \cite{yue2024mmmu}. MIRB and MuirBench more specifically test multi-image perception, comparison, retrieval, and multi-hop reasoning \cite{zhao2024mirb,wang2025muirbench}, while MANTIS studies interleaved multi-image instruction tuning \cite{jiang2024mantis}. OSEF separates evidence selection from action planning in cross-video procedure planning \cite{ye2026osef}. These studies primarily measure end-to-end capability; we instead intervene on intermediate information to identify where a controlled failure becomes recoverable.

\subsection{Long-Context Retrieval and Position Sensitivity}

Language models often use long contexts non-uniformly, especially when relevant information appears near the middle \cite{liu2024lost}. MM-NIAH, MMNeedle, and Visual Haystacks extend this concern to multimodal retrieval and cross-image reasoning \cite{wang2024mmniah,wang2025mmneedle,wu2025visualhaystacks}. Our task likewise embeds two relevant sources among distractors, but additionally requires their composition through an item-specific rule.

Related systems manage evidence through adaptive selection, graph traversal, compression, or multi-agent retrieval \cite{xiong2026adaptive,zhang2026bridging,gao2026dspc,lou2026racer}. Work on adaptive temporal context and observation windows similarly shows that useful evidence depends on task structure and context budget \cite{duan2025adaptive,han2026earlyearlyenoughdesigndependent}. Provenance-aware memory and worksheet-level retrieval preserve evidence units that might otherwise be fragmented \cite{ProvenanceMemoryRetrieval,SheetAsToken}. These mechanisms are outside our experiment, but they motivate treating evidence recovery as a separable stage.

Input order can change reasoning even when semantics are preserved \cite{chen2024premise}. Modality order and multimodal perturbations affect VLM predictions \cite{wardle2024imagefirst,wei2026order}, while prompt privilege captures related variation across semantically equivalent requests \cite{jin2026same}. We therefore balance evidence order and absolute position so that placement does not determine the main Oracle contrast.

\subsection{Oracle and Decomposition Diagnostics}

DISSECT uses Human and Model Oracle conditions to separate visual extraction, perception, and integration failures in scientific VQA \cite{kukreja2026dissect}. Perceptual Taxonomy distinguishes recognition from cross-property reasoning \cite{lee2025perceptual}, and replacing puzzle images with text has similarly isolated perception as a limiting stage \cite{sharif2026disentangling}. Question decomposition can then expose and selectively repair intermediate VQA steps \cite{khan2023decomposition}. These studies are the closest methodological basis for our design.

Process-oriented language-model work uses mid-level supervision, reasoning-trace decomposition, controlled access to memorized information, or staged reading to expose hidden failure points \cite{jiang2026scribestructuredmidlevelsupervision,jiang-etal-2026-drp,jiang-ferraro-2026-beyond,han2026read}. Traceability, provenance, dependency-guided rollback, and epistemic-source analysis provide related ways to audit or repair intermediate state \cite{AgentTracesToTrust,IndustrialVLA-Bench,RollbackRepair,EpistemicAdmission}.

More distant analogues use matched controls to isolate gains from model fusion and encoder choice \cite{han2026validationstagecombinatorialfusionanalysis,han2026interpretablevslearnedencoders}. Factor decomposition has also been applied to markets, crisis responses, broadband disparities, and rare-association recovery \cite{han2026interpretablefactordecompositiondecision,han2026decomposingfirmlevelcrisisresponses,han2026explainablemachinelearningbroadband,han2026supportthresholdsalgorithmslimit}. Staged interventions in traffic control, trajectory prediction, and agent collaboration share the broader idea of identifying unreliable intermediate components \cite{duan2025bayesian,duan2026maven,lou2026drf}. Workflow root-cause analysis provides a domain-level analogue of localizing failure in a structured pipeline \cite{li2022root}.

Our design differs from these approaches in its paired intervention grain. Within each item, it independently supplies two intermediate information sources while retaining the same 16-image context, question, and answer across a complete $2\times2$ design. The comparison therefore asks whether recovery requires repairing the codebook or the ordered target instances.

\section{Factorial Oracle Design and Task Instantiation}

This section first defines the diagnostic intervention and then instantiates it in the 16-image task. Following the general logic of Oracle-based diagnosis \cite{kukreja2026dissect} and selective intermediate repair \cite{khan2023decomposition}, the design holds the original context and question fixed while independently repairing two candidate intermediate representations. It is an evaluation design for locating a recoverable deficit, rather than a new inference-time solver.

\subsection{Task Pipeline}

Each item requires the following sequence of operations:

\begin{enumerate}
    \item locate two characters by the objects they hold;
    \item identify each character's color and head shape;
    \item preserve the requested target order;
    \item retrieve two color codes and two shape codes from a visual schema;
    \item compose and output two ordered code names.
\end{enumerate}

\begin{figure*}[t]
    \centering
    \includegraphics[width=\textwidth]{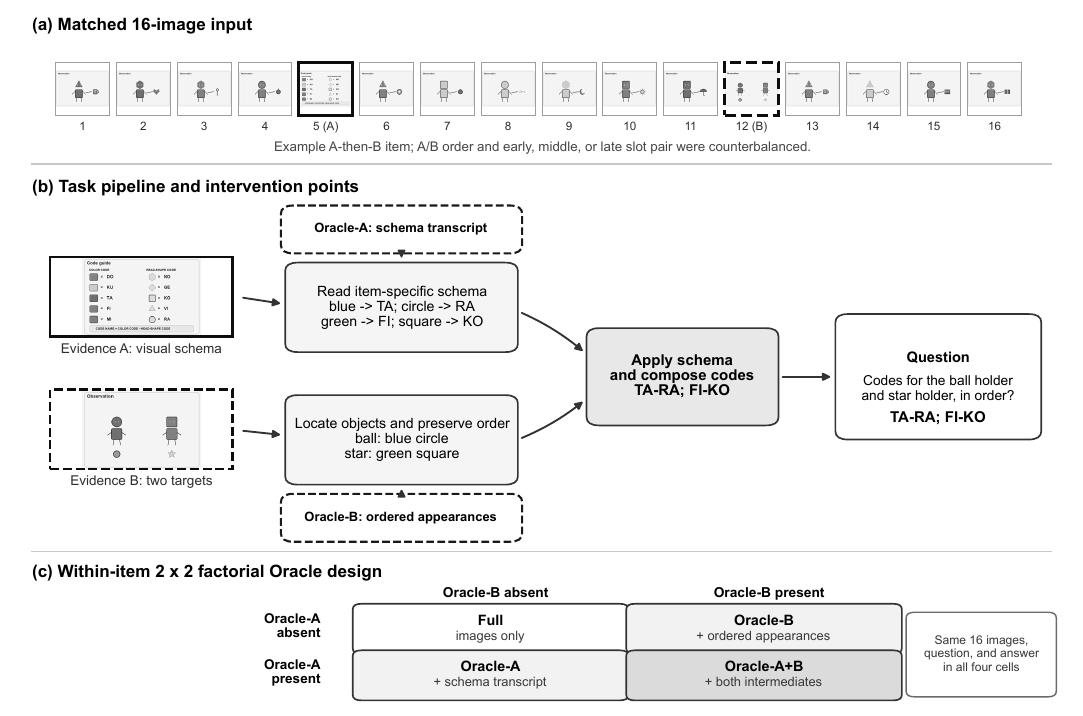}
    \caption{Overview of the multi-image schema-application task. (a) One item contains Evidence A, Evidence B, and 14 fixed distractors; A/B order and slot pair are counterbalanced. (b) Oracle-A supplies the item-specific schema transcript, whereas Oracle-B supplies the ordered target appearances. Both intermediate representations feed the same code-composition step. (c) Crossing the interventions gives four within-item conditions with identical images, question, and expected answer. Task-relevant color identities are also named in text}
    \label{fig:task}
\end{figure*}

Fig.~\ref{fig:task} summarizes one matched item, the two task-relevant information paths, and the factorial intervention points.

\subsection{Visual Schema A}

Evidence A is a visual codebook containing five colors (red, blue, green, yellow, and purple), five head shapes (circle, square, triangle, hexagon, and diamond), and the rule \texttt{COLOR CODE - HEAD-SHAPE CODE}. Ten distinct two-letter nonce tokens are sampled without replacement for each item and assigned independently to the five colors and five shapes. Mapping rows are visually randomized. Because mappings are regenerated for every item, no code token has a stable meaning across items.

\subsection{Instance Evidence B and Distractors}

Evidence B contains two characters holding different target objects. Active characters use four colors (red, blue, green, and yellow) and four head shapes (circle, square, triangle, and hexagon). The two targets differ on both attributes. Fourteen distractor images instantiate the remaining 14 combinations exactly once. Distractor objects are unique, and neither target object appears outside Evidence B. Purple and diamond are schema-level distractor mappings and do not describe target or distractor characters.

The complete sequence therefore contains one schema image, one dual-target image, and 14 distractor images. Evidence A and B occupy slots $(2,9)$, $(5,12)$, or $(8,15)$, covering early, middle, and late centers while maintaining a seven-slot separation. Half of the items present A before B and half present B before A. Position and evidence order are balanced nuisance variables.

\subsection{Main Conditions}

Every item is evaluated in the four conditions shown in Table~\ref{tab:conditions}. Within an item, the ordered image list, final question, expected answer, response instruction, and model settings are held constant. Only the verified-information fields change.

\begin{table}[t]
\centering
\caption{Within-item factorial conditions}
\label{tab:conditions}
\setlength{\tabcolsep}{4pt}
\begin{tabular}{lccc}
\toprule
Condition & 16 images & Schema text & Instance text \\
\midrule
Full ($A0B0$)       & Yes & No  & No  \\
Oracle-A ($A1B0$)   & Yes & Yes & No  \\
Oracle-B ($A0B1$)   & Yes & No  & Yes \\
Oracle-A+B ($A1B1$) & Yes & Yes & Yes \\
\bottomrule
\end{tabular}
\end{table}

\oracleA{} supplies all five color mappings, all five shape mappings, and the composition rule as a canonical text transcript. It does not mark answer-relevant rows, mention target appearances, or include a final code name. The model must still locate and recognize both targets from the images.

\oracleB{} supplies two fixed-template sentences describing the target characters in the requested order, for example, ``The character holding the cup is blue and has a triangle head.'' It contains no schema token or final answer. The model must still read the item-specific visual schema, retrieve four nonce codes, compose two names, and output them in order. \oracleB{} repairs the whole instance side of the task at once: target-object localization, character selection, visual-attribute recognition, within-character binding, and target-order preservation. We call this an \emph{ordered instance-information intervention}.

\subsection{Auxiliary Controls}

Three auxiliary diagnostics run on the same balanced 36-item subset. The \oracleAB{} text-only ceiling removes all images while providing the full schema and target appearances. Retrieval-A retains all 16 images, provides the target appearances, and asks for their code names, isolating visual-schema use. Retrieval-B keeps the same 16-image sequence and asks directly for the two ordered appearances. Its instruction tells the model to ignore the code guide, avoid two-letter codes, use only the four active colors and shapes, and answer as \texttt{COLOR SHAPE; COLOR SHAPE}. These diagnostics are not treated as another factorial test; they are there to show what the main rescue pattern is made of.

\section{Experimental Design and Evaluation Protocol}

The experiment is organized to estimate diagnostic repair contrasts rather than to train or rank models. Dataset balance, fixed inference settings, randomized run order, paired outcomes, and item-cluster inference are used so that condition differences can be attributed to the supplied intermediate information rather than to item composition, presentation order, or decoding settings.

\subsection{Dataset and Balance}

The dataset contains 72 newly generated items. Table~\ref{tab:balance} summarizes the balance constraints. Every active appearance occurs nine times across the 144 target roles, and every target-appearance pair is unique. The same balanced 36-item subset is used for all auxiliary controls.

\begin{table}[t]
\centering
\caption{Dataset balance}
\label{tab:balance}
\begin{tabular}{lr}
\toprule
Design property & Count \\
\midrule
Base items & 72 \\
Target roles & 144 \\
Occurrences per active appearance & 9 \\
Occurrences per target object & 24 \\
Items per evidence center & 24 \\
A-before-B / B-before-A items & 36 / 36 \\
Items per center-by-order cell & 12 \\
Auxiliary subset & 36 \\
\bottomrule
\end{tabular}
\end{table}

\subsection{Models and Inference Settings}

We evaluate the fixed panel in Table~\ref{tab:models}. The two OpenAI models provide a same-family full-versus-mini comparison, while Claude Sonnet 5 provides cross-provider replication.

\begin{table}[t]
\centering
\caption{Model panel and inference settings}
\label{tab:models}
\setlength{\tabcolsep}{3pt}
\begin{tabularx}{\columnwidth}{lXl}
\toprule
Provider & Model identifier & Setting \\
\midrule
OpenAI & \texttt{gpt-4.1-2025-04-14} & temp. 0 \\
Anthropic & \texttt{claude-sonnet-5} & thinking off \\
OpenAI & \texttt{gpt-4.1-mini-2025-04-14} & temp. 0 \\
\bottomrule
\end{tabularx}
\end{table}

Every observation is an independent API request with no conversation history. OpenAI images use high detail; Anthropic extended thinking is disabled. We evaluate fixed non-deliberative API configurations rather than each model's maximum attainable performance. The prompt requests a final line formatted as \texttt{Final answer: <answer>}, and no chain-of-thought is requested.

The analytic set contains 288 main observations per model ($72\times4$) and 108 auxiliary observations per model ($36\times3$), yielding 396 observations per model and 1,188 analyzed responses.

\subsection{Run Order}

The four main conditions were randomized in blocks. Each block contained every item once before any item recurred in another condition, and adjacent requests never shared an item. The same randomized schedule was used across the three models to preserve item/condition balance. Retrieval-B used a separate randomized order over the 36-item auxiliary subset.

\subsection{Outcomes}

The primary outcome is strict dual-target exact accuracy: both code names, target order, and color/shape code order must be semantically correct, and the response must use the requested \texttt{Final answer:} wrapper. Missing or malformed final-answer lines count as incorrect. We also report semantic accuracy without the wrapper requirement, target-level exact accuracy, four-atom code accuracy, exact target-order reversal, and Retrieval-B partial-credit scores.

\subsection{Statistical Analysis}

Let $p_{00}$, $p_{10}$, $p_{01}$, and $p_{11}$ denote exact accuracy in Full, \oracleA{}, \oracleB{}, and \oracleAB{}, respectively. For model $m$, ordered instance-information rescue is $p_{01,m}-p_{00,m}$. The primary panel estimand gives equal weight to the three fixed models:

\begin{equation*}
\Delta_{B}=\frac{1}{3}\sum_{m=1}^{3}\left(p_{01,m}-p_{00,m}\right).
\end{equation*}

The second panel estimand measures rescue asymmetry:

\begin{equation*}
\Delta_{B-A}=\frac{1}{3}\sum_{m=1}^{3}\left(p_{01,m}-p_{10,m}\right).
\end{equation*}

The panel estimate gives equal weight to the three models. We compute 95\% confidence intervals using 10,000 item-cluster bootstrap draws; each draw retains every model and condition observation for a sampled item. One-sided item-cluster sign-flip tests use 10,000 draws, with Holm correction across the two primary tests. Model-specific \oracleB{} effects are reported with confidence intervals to show whether the pattern repeats by model.

Secondary quantities include schema rescue $p_{10}-p_{00}$, conditional schema rescue $p_{11}-p_{01}$, conditional instance rescue $p_{11}-p_{10}$, joint rescue $p_{11}-p_{00}$, factorial interaction $p_{11}-p_{10}-p_{01}+p_{00}$, and post-Oracle residual $1-p_{11}$. We report all contrasts directly on the probability scale. Robustness analyses compare semantic accuracy with strict format compliance and stratify the two main contrasts by evidence center and order.

\section{Results}

The results follow the claimed contributions. Primary factorial contrasts identify which repair restores the task; within-item transitions and representation controls test that localization; and position and order strata test its robustness. These quantities measure diagnostic recoverability, not superiority over an existing inference system.

\subsection{Primary Oracle Effects}

\begin{figure*}[t]
    \centering
    \includegraphics[width=\textwidth]{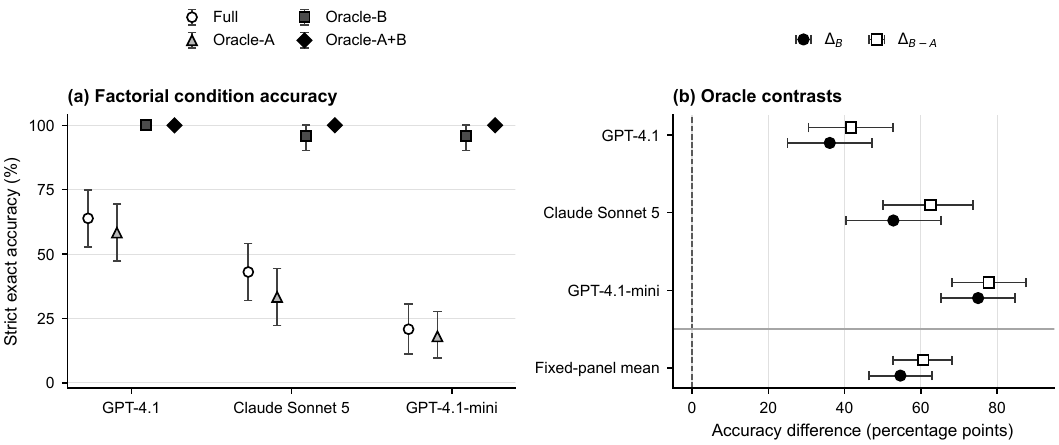}
    \caption{Factorial Oracle results under strict exact scoring. (a) Condition accuracy by model; error bars are 95\% item-bootstrap intervals. (b) Oracle-B rescue ($\Delta_B$) and the Oracle-B versus Oracle-A asymmetry ($\Delta_{B-A}$) by model and for the equal-weighted fixed panel; horizontal bars are 95\% item-cluster bootstrap intervals}
    \label{fig:results}
\end{figure*}

The Oracle contrast is large and one-sided (Table~\ref{tab:primary-contrasts}). Supplying verified ordered instance information increased strict exact accuracy by 54.6 percentage points across the equal-weighted fixed model panel (95\% item-cluster bootstrap CI: 46.3--63.0). Oracle-B accuracy exceeded Oracle-A accuracy by 60.6 points (95\% CI: 52.8--68.1). The one-sided item-cluster sign-flip tests yielded $p\leq .0001$ for each contrast, and both remained significant after Holm correction ($p_{\mathrm{Holm}}\leq .0002$). Fig.~\ref{fig:results} shows the corresponding model-specific condition means and contrasts.

\begin{table}[t]
\centering
\caption{Primary fixed-panel Oracle contrasts}
\label{tab:primary-contrasts}
\setlength{\tabcolsep}{3pt}
\begin{tabular}{lcccc}
\toprule
Contrast & Est. & 95\% CI & Raw $p$ & Holm $p$ \\
\midrule
$\Delta_B$ & +54.6 pp & [46.3, 63.0] & $\leq .0001$ & $\leq .0002$ \\
$\Delta_{B-A}$ & +60.6 pp & [52.8, 68.1] & $\leq .0001$ & $\leq .0002$ \\
\bottomrule
\end{tabular}
\end{table}

The same pattern appears in every model. Oracle-B rescue was +36.1 points for GPT-4.1 (95\% CI: 25.0--47.2), +52.8 for Claude Sonnet 5 (40.3--65.3), and +75.0 for GPT-4.1-mini (65.3--84.7). The weaker the unassisted model, the larger the absolute rescue, but the direction never changed.

\subsection{Factorial Profiles and Residual Error}

Table~\ref{tab:model-results} reports the full factorial profile and residual error by model.

\begin{table*}[t]
\centering
\caption{Strict exact accuracy and probability-scale Oracle effects by model}
\label{tab:model-results}
\renewcommand{\arraystretch}{1.08}
\begin{tabular*}{\textwidth}{@{\extracolsep{\fill}}lrrrrrrrr@{}}
\toprule
\multirow{2}{*}{Model}
& \multicolumn{4}{c}{Strict exact accuracy (\%)}
& \multicolumn{3}{c}{Oracle contrasts (pp)}
& \multirow{2}{*}{Residual (\%)} \\
\cmidrule(lr){2-5}\cmidrule(lr){6-8}
& Full & O-A & O-B & O-AB & B rescue & A rescue & $A\!\times\!B$ & \\
\midrule
GPT-4.1 & 63.9 & 58.3 & 100.0 & 100.0 & +36.1 & $-$5.6 & +5.6 & 0.0 \\
Claude Sonnet 5 & 43.1 & 33.3 & 95.8 & 100.0 & +52.8 & $-$9.7 & +13.9 & 0.0 \\
GPT-4.1-mini & 20.8 & 18.1 & 95.8 & 100.0 & +75.0 & $-$2.8 & +6.9 & 0.0 \\
\bottomrule
\end{tabular*}
\par\smallskip
\parbox{\textwidth}{\footnotesize\emph{Note}: O-A, O-B, and O-AB denote Oracle-A, Oracle-B, and Oracle-A+B. Rescue and interaction entries are percentage points; residual is $100(1-p_{11})$.}
\end{table*}

Full-task accuracy ranged from 20.8\% to 63.9\%, but Oracle-B raised every model to at least 95.8\%, and Oracle-A+B raised all models to 100\%. Averaged across models, strict Oracle-A rescue was $-6.0$ points, joint rescue was +57.4 points, conditional schema rescue after Oracle-B was +2.8 points, and the factorial interaction was +8.8 points. The zero observed post-Oracle residual and 100\% text-only ceiling leave little room for a downstream explanation: when both the schema and the ordered appearances were available, lookup, composition, ordering, and output all worked.

Strict scoring also captured instruction-following failures, especially for Claude and GPT-4.1-mini. GPT-4.1 complied on 100\% of responses in every main cell. Claude's Full, Oracle-A, Oracle-B, and Oracle-A+B compliance rates were 88.9\%, 50.0\%, 97.2\%, and 100\%; GPT-4.1-mini's were 79.2\%, 58.3\%, 100\%, and 100\%. Semantic-only scoring gives the same substantive answer (Table~\ref{tab:semantic}): Oracle-B rescue remained +52.3 points (95\% CI: 44.0--60.2), and Oracle-B still exceeded Oracle-A by +51.9 points (43.5--60.2). The strict Oracle-A drop was mostly a formatting artifact; semantically, schema text alone barely changed accuracy (+0.5 points).

\begin{table*}[t]
\centering
\caption{Semantic-only sensitivity analysis (percent)}
\label{tab:semantic}
\setlength{\tabcolsep}{5pt}
\begin{tabular}{lrrrrrr}
\toprule
Model & Full & O-A & O-B & O-AB & $\Delta_B$ & $\Delta_{B-A}$ \\
\midrule
GPT-4.1 & 63.9 & 58.3 & 100.0 & 100.0 & +36.1 & +41.7 \\
Claude Sonnet 5 & 48.6 & 52.8 & 98.6 & 100.0 & +50.0 & +45.8 \\
GPT-4.1-mini & 25.0 & 27.8 & 95.8 & 100.0 & +70.8 & +68.1 \\
\midrule
Fixed-panel mean & 45.8 & 46.3 & 98.1 & 100.0 & +52.3 & +51.9 \\
\bottomrule
\end{tabular}
\end{table*}

\subsection{Within-Item Rescue Transitions}

\begin{figure*}[t]
    \centering
    \includegraphics[width=\textwidth]{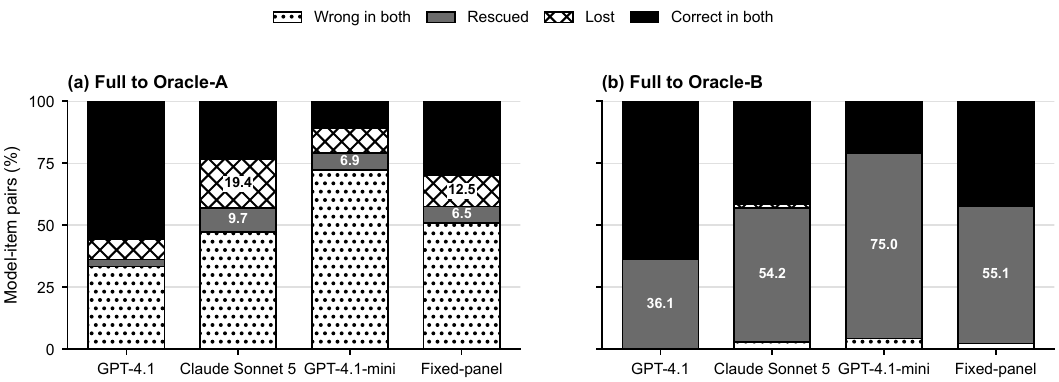}
    \caption{Descriptive within-item transitions from Full to each single-Oracle condition under strict exact scoring. Each bar partitions the same model--item pairs into persistently wrong, rescued, lost, and persistently correct outcomes. The fixed-panel bar aggregates three equally sized model samples. Numbers inside segments are percentages and emphasize rescue and larger loss components}
    \label{fig:transitions}
\end{figure*}

The within-item transitions show the same asymmetry (Fig.~\ref{fig:transitions}). Oracle-A rescued 14 Full failures but lost 27 Full successes; 110 pairs remained wrong in both conditions. Oracle-B rescued 119 Full failures, lost only one Full success, and left only five pairs wrong in both conditions. Under semantic scoring, Oracle-A rescued 19 and lost 18, while Oracle-B rescued 113, lost none, and left four wrong in both. Oracle-B was not trading one error pattern for another. It was recovering the same items that failed without instance information.

\subsection{Representation-Level Diagnostics}

\begin{figure*}[t]
    \centering
    \includegraphics[width=\textwidth]{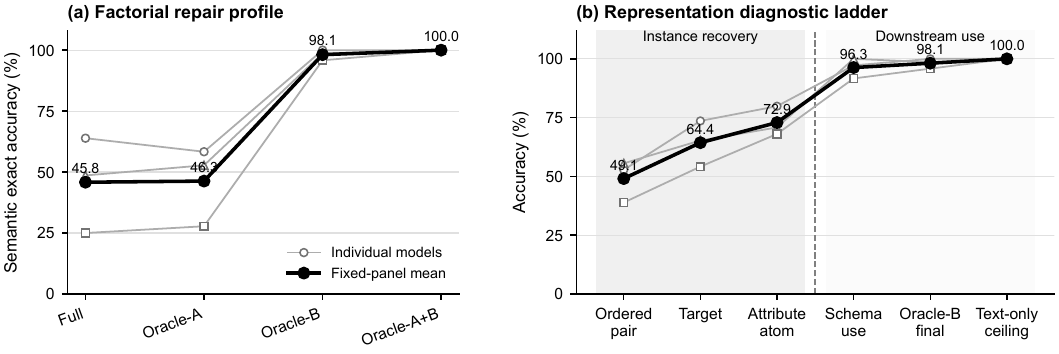}
    \caption{Representation-level diagnostic summary. Thin gray lines show individual models, and the thick black line shows the equal-weighted fixed-panel mean. (a) Semantic final-task accuracy across the four factorial conditions. (b) Diagnostic ladder separating ordered instance recovery from downstream schema use and composition. Retrieval-B scores the ordered target appearance at pair, target, and attribute-atom levels; Retrieval-A scores visual-schema use when target appearances are provided}
    \label{fig:representation}
\end{figure*}

The combined repair pattern appears in Fig.~\ref{fig:representation}. In the final-task profile, the three models follow the same trajectory: Oracle-A leaves semantic accuracy nearly unchanged, while Oracle-B raises performance close to ceiling. On the balanced 36-item auxiliary subset, Oracle-A+B text-only accuracy was 36/36 for every model. Retrieval-A was also high under semantic scoring: 97.2\% for GPT-4.1, 100.0\% for Claude Sonnet 5, and 91.7\% for GPT-4.1-mini. The visual schema was therefore usually usable once the ordered target appearances were already specified.

Retrieval-B produced the opposite profile. Ordered appearance exact accuracy was 52.8\% for GPT-4.1, 55.6\% for Claude Sonnet 5, and 38.9\% for GPT-4.1-mini, for a fixed-panel mean of 49.1\% (95\% item-cluster bootstrap CI: 36.1--62.0). No final answer contained a nonce code, so the low score was not caused by confusing appearance outputs with code-name outputs. The models often failed to recover the appearances themselves.

Retrieval-B errors were often partial rather than wholesale. Average target-level accuracy was 64.4\%, and individual color/shape atom accuracy was 72.9\%, compared with 49.1\% for the complete ordered pair. Among the 50 incorrect but canonical Retrieval-B responses, 21 differed from the gold answer in one of four attribute atoms, 18 differed in two atoms, four differed in three, and seven differed in all four; 33 still preserved one complete target appearance. No response exactly reversed two otherwise correct target appearances. Retrieval-B success also tracked Full-task success for the same item and model: Full semantic accuracy was 77.4\% when Retrieval-B was correct and 23.6\% when it was incorrect.

\subsection{Evidence Position and Order}

The same direction appears across the nuisance strata. Oracle-B rescue ranged from +51.4 to +58.3 points across early, middle, and late evidence centers. It was positive for both evidence orders and larger when B preceded A (+63.0 points) than when A preceded B (+46.3). Oracle-B exceeded Oracle-A in every stratum.

\section{Discussion}

This section interprets the findings against the three gaps identified in Section II: end-to-end multi-image evaluation, evidence use in long contexts, and diagnosis at intermediate task boundaries. It also states the scope of the methodological and empirical contributions.

\subsection{Where the Task Breaks}

The main experiment gives a direct answer to the paper title. End-to-end strict accuracy averaged 42.6\% across the fixed panel. Supplying the complete schema alone barely changed semantic accuracy (+0.5 points). Supplying the ordered target appearances raised semantic accuracy by 52.3 points and strict accuracy by 54.6 points. With both verified sources, all three models reached 100\%. The dominant failure is on the instance side of the task: the models often lose the ordered target appearances, then perform well once that information is handed to them.

Oracle-B did not give away the final answer. It gave colors and shapes, but no item-specific schema tokens and no code names. The model still had to read the visual codebook, execute four nonce-code lookups, compose two names, and preserve order. Those downstream operations were not the limiting step in this experiment. The missing piece was the ordered instance representation.

\subsection{Converging Evidence for an Instance-Information Bottleneck}

The supporting evidence is consistent across analyses. Complete schema transcription changed semantic panel accuracy by only +0.5 points, while verified ordered target appearances produced a +52.3-point rescue. In strict within-item transitions, Oracle-B rescued 119 failures and lost only one success; Oracle-A rescued 14 failures but lost 27 successes. Retrieval-A was high under semantic scoring (91.7--100\%), so the visual schema was usually usable once appearances were specified. Retrieval-B, by contrast, recovered the complete ordered appearances on only 49.1\% of observations, even though atom-level accuracy was higher (72.9\%) and exact pair reversals were absent. Full semantic accuracy was 77.4\% when Retrieval-B succeeded and only 23.6\% when it failed. These results make the bottleneck hard to place downstream. The main recoverable deficit is the construction of the ordered target-instance representation.

The comparison does not require schema acquisition and target retrieval to be equally hard. It asks which repaired intermediate representation restores the same final task, and the answer is consistently the ordered instance representation. The experiment fixes sequence length at 16 images, so it does not separate the intrinsic complexity of Evidence B from the burden added by the distractors. Within this 16-image schema-application setting, however, the result is decisive: repairing the instance representation restores the task; repairing the schema does not.

\subsection{Implications for Multi-Image Evaluation}

Broad multi-image benchmarks such as MIRB characterize task-level capability \cite{zhao2024mirb}; the present design instead measures recoverability at predefined intermediate boundaries. DISSECT uses several Oracle modes to partition scientific VQA failures \cite{kukreja2026dissect}; our narrower contribution is a matched $2\times2$ contrast that estimates independent and joint repairs while preserving the visual context. The design therefore complements benchmark and inference methods rather than claiming that a new solver outperforms them.

This distinction matters for system design: improving visual retrieval, external grounding, or intermediate-state verification may be more effective than adding further symbolic reasoning scaffolds when the dominant error occurs upstream. At the systems level, CE-Agent routes uncertain multi-step agent states from an edge model to a cloud model and distills successful offloaded trajectories back to the edge \cite{huang2026agent}. The bottleneck identified here suggests that ordered instance recovery is a candidate signal for similarly targeted verification or escalation in multimodal systems.

The three models differed substantially in unassisted accuracy, but the condition pattern was the same: Oracle-A produced little semantic gain, Oracle-B produced a large rescue, and Oracle-A+B reached 100\%. GPT-4.1 had the highest Full accuracy and the smallest absolute rescue; GPT-4.1-mini had the lowest Full accuracy and the largest rescue; Claude Sonnet 5 lay between them.

\section{Limitations}

The claimed scope is deliberately narrow: the design localizes a recoverable bottleneck within one controlled task, two coarse intermediate representations, and a fixed model panel.

The experiment uses synthetic images rather than natural photographs or documents. This makes the colors, shapes, objects, and nonce codes exactly controllable, but it also narrows the empirical setting. The next step is to test whether the same rescue profile appears with natural images, document-like layouts, video frames, and more realistic multimodal workflows.

The Oracle treatments also arrive as text immediately before the question, so their effects combine semantic correction with textual accessibility and recency. A future version could add matched irrelevant text or delayed Oracle information to separate these components more directly. In the present design, making the ordered appearances explicit was sufficient for the models to finish the task.

\oracleB{} repairs several upstream operations at once: target localization, character selection, attribute recognition, binding, and target-order preservation. This study identifies the larger failing region, not the smallest failing subroutine inside it. A more granular follow-up could separate these subprocesses by supplying the target object locations, the selected character identities, the unordered attributes, or the target order in separate Oracle cells. Retrieval-B suggests that errors accumulate across local selection and binding decisions, but those finer interventions are needed to isolate which upstream subprocesses dominate.

Strict exact accuracy mixes answer correctness with instruction following. That mattered for Claude Sonnet 5 and GPT-4.1-mini, especially in Oracle-A. The semantic-only analysis keeps the main claim intact: schema text alone barely helps, while ordered target appearances produce a large rescue.

The model panel is small and proprietary, and hosted model versions change. Replication on newer and open-weight models is still necessary. The design also fixes the context at 16 images and uses purple and diamond only as schema distractors. It does not map how the bottleneck grows with sequence length, distractor count, or visual similarity.

\section{Conclusion}

This paper presents a within-item factorial Oracle design for diagnosing failures in multi-image visual schema application. By preserving the original visual context and final question while selectively supplying verified schema and ordered instance information, the design turns aggregate accuracy into a repair test. Verified ordered instance information rescued strict accuracy by 54.6 percentage points, and Oracle-B exceeded Oracle-A by 60.6 points across the model panel. The effect appeared in all three models, survived semantic-only scoring, and reached a 100\% joint-Oracle ceiling. At the same time, Retrieval-B recovered the complete ordered appearances on only 49.1\% of observations. The tested models usually completed the downstream schema operation once the required visual facts were supplied; the harder step was extracting those facts from the image sequence.

\end{document}